# Warping Peirce Quincuncial Panoramas

Chamberlain Fong [spectralfft@yahoo.com],   Brian Vogel

**Abstract**

The Peirce quincuncial projection is a mapping of the surface of a sphere to the interior of a square. It is a conformal map except for four points on the equator. These points of non-conformality cause significant artifacts in photographic applications. In this paper, we propose an algorithm and user-interface to mitigate these artifacts. Moreover, in order to facilitate an interactive user-interface, we present a fast algorithm for calculating the Peirce quincuncial projection of spherical imagery. We then promote the Peirce quincuncial projection as a viable alternative to the more popular stereographic projection in some scenarios.

## 1  Introduction

The Peirce quincuncial projection[1][2] is a mapping of the surface of a sphere to a square. It was originally proposed by Charles Sanders Peirce in 1879 for geographic applications. Indeed, Peirce was working for U.S. Coast and Geodetic Survey when he came up with this projection. In recent years, the fisheye photographic community[3] has adopted this projection for aesthetic purposes. The figure below shows Peirce's projection of the world and an example of its use for photographic purposes.

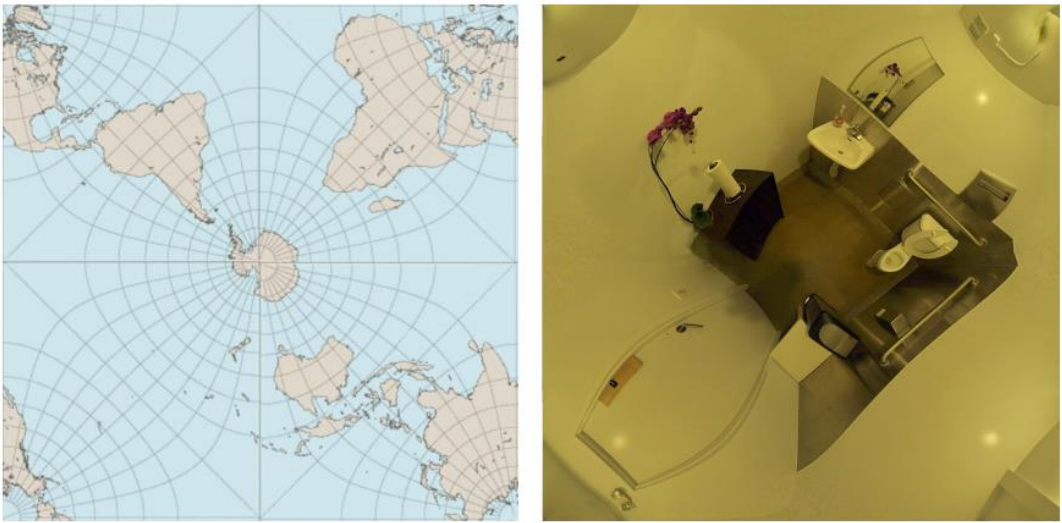

Figure 1: Peirce's quincuncial projection of the world and a photographic example

## 2 Spherical Panoramas

Anyone who has used street view in Google maps should be familiar with these types of photographs. There is a sphere of light that shines upon us all the time. Advances in photography and computer graphics[4][5] has allowed the photographer to capture and view this sphere. These type of photographs are called *spherical panoramas*. They can either be viewed interactively using a computer as is the case with Google street view; or using some sort of projection mapping the surface of the sphere to a planar image. There are several photographers who specialize on spherical panoramas.

There is a theorem in differential geometry that states that the sphere is not a developable surface. This means that it cannot be flattened onto a plane without distortion. Since the time of Ptolemy circa 100 A.D., geographers and cartographers have come up with various projections of the sphere to the plane. This is a significant problem in geography and map-making that there are whole books[6] devoted to this topic. Every type of spherical projection has some sort of compromise in which it distorts distance, area, scale, direction, or angles in order to represent the sphere on a plane.

The spherical panorama community[7] has adopted several spherical projection techniques to view their photographs. The standard projection used by spherical panorama photographers is called the *equirectangular* projection. Most spherical panoramas are stored and viewed as equirectangular images. Equirectangular images have an image dimension ratio of 2:1. This follows from the fact that the sphere has 360° in longitude and 180° in latitude. An example of an equirectangular image is show below. We have color-coded some important parts for illustration purposes. The equator is colored red. The north and south poles of the sphere are colored purple and yellow, respectively. In spherical panorama jargon, the north pole is known as the *zenith* and the south pole is known as the *nadir*.

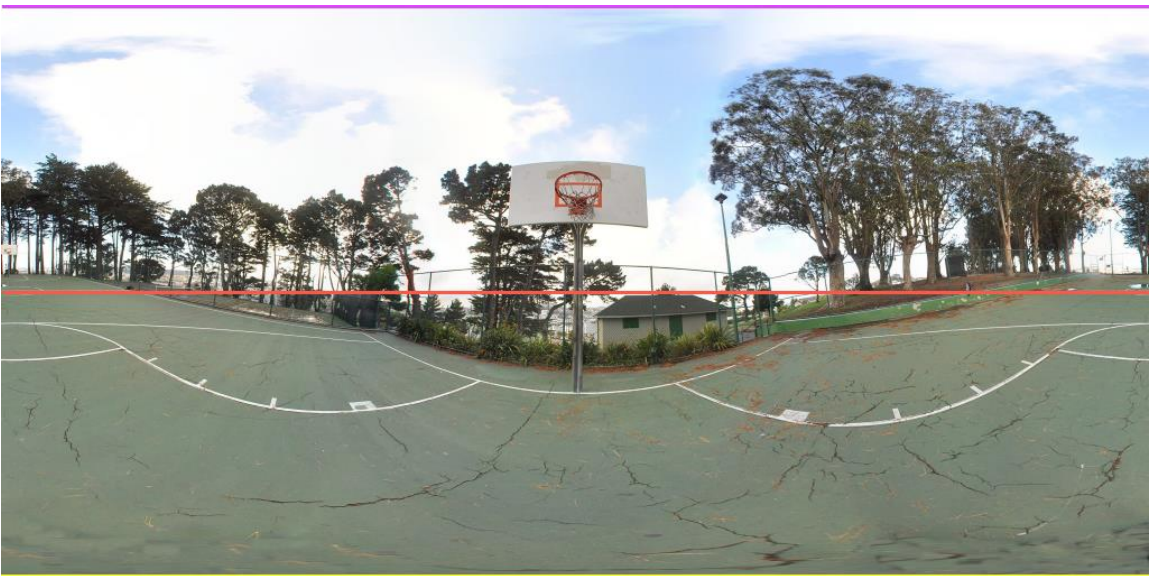

Figure 2: An equirectangular panorama

Another popular projection among spherical panorama photographers is the *stereographic* projection. The stereographic projection is conformal, which means that it preserves angles locally within the image. This property makes stereographic panoramas aesthetically pleasing to the eyes. In fact, there is a thriving online community[8] of photographers who use the stereographic projection to make artful "little planets" that simulate an overhead aerial view. Figure 3 shows an example of the stereographic projection applied to our equirectangular image in figure 2. There are two types of stereographic projections depending on which pole the projection occurs.

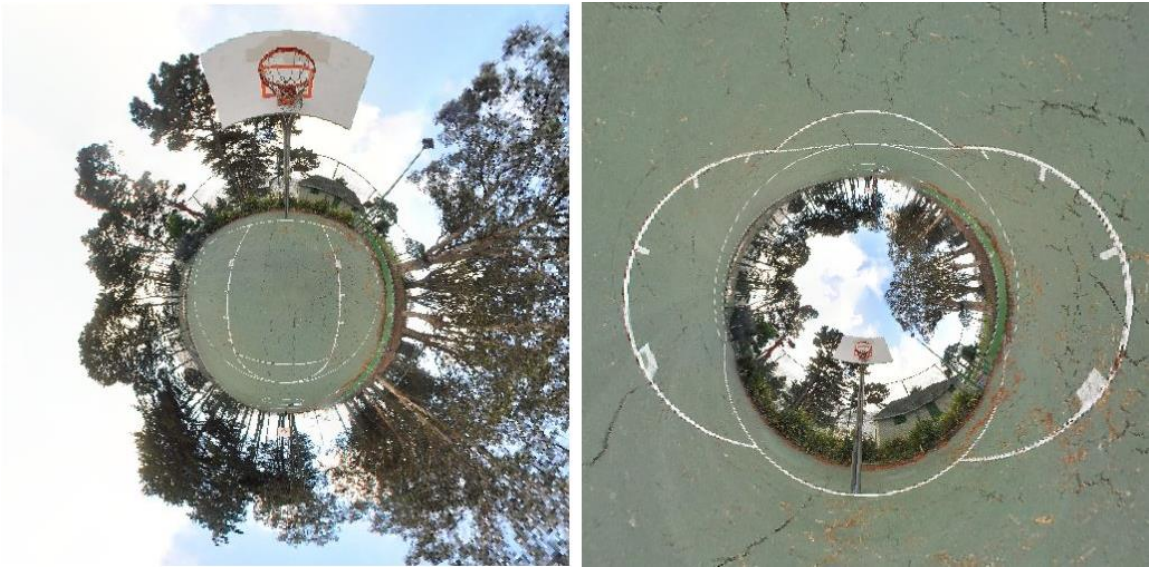

Figure 3: Stereographic panorama -- zenith type (left) & nadir type (right)

3  **The Peirce Quincuncial Projection**

The Peirce quincuncial projection is closely related to the stereographic projection. It was devised by Peirce after drawing inspiration from Hermann Schwarz's mathematical result on a conformal mapping of the circle onto a polygon with n sides. This result is called the *Schwarz-Christoffel transformation* and is an important theorem in the field of Complex Analysis.

The Peirce quincuncial projection has several features that make it visually interesting. One pole lies exactly in the center of the square, while the other pole is split among the four corners of the square. The equator lies on an inner square that is formed by connecting the midpoints of each of the edges in the perimeter of the square. This inner square is tilted by 45° when compared to the outer square.

The Peirce quincuncial projection is a conformal mapping, except for four points lying on the equator. Conformality is important for aesthetic reasons in photographic applications. We will talk more about the non-conformal points in a latter section of this paper.

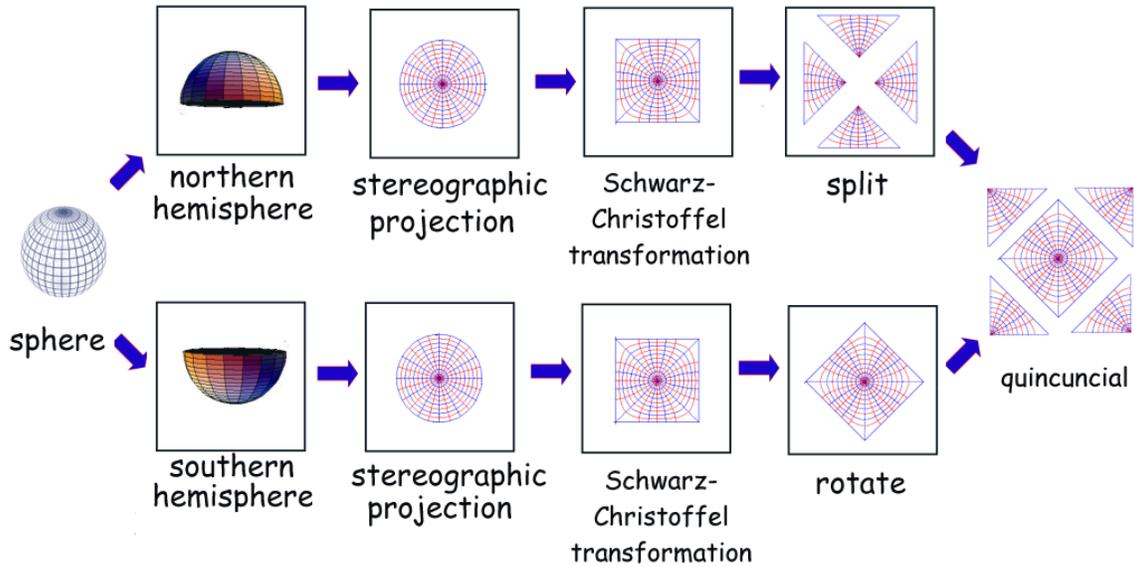

Figure 4: Overview of the Peirce quincuncial projection

An overview of the Peirce Quincuncial projection is shown in figure 4. In this figure, we start from a sphere and produce a square projection. The first step splits the sphere into northern and southern hemispheres. The next step does a stereographic projection on the two hemispheres. This is followed by using the Schwarz-Christoffel transformation on these two stereographic discs to produce two squares. These two squares are then reassembled as a larger square in a quincuncial fashion.

The main equation provided by Peirce for his quincuncial projection is:

$$\tan\left(\frac{p}{2}\right) e^{i\theta} = \text{cn}\left(z, \frac{1}{\sqrt{2}}\right)$$

where $w = pe^{i\theta}$ and $z = x + iy$

In this equation, every point in the stereographic unit disk represented by the complex variable *w* with polar coordinates (p,θ) is mapped to complex variable *z* in the Peirce square with (x,y) coordinates. For imaging applications, it suffices to know that Jacobi elliptic function cn(*z*,1/√2) can be used to map the circle to a square in the complex plane conformally, and vice versa. Note that the complex Jacobi elliptic function is just a special case of the more general Schwarz-Christoffel transformation. See Appendix A for sample Matlab source code of the Peirce Quincuncial projection for imaging applications.

Figure 5 shows a detailed example of the Peirce quincuncial projection steps applied to a real spherical panorama. We start with our equirectangular image split into its northern and southern hemispheres and proceed with intermediate images corresponding to steps in the Peirce quincuncial projection. The projection is called quincuncial because the two small squares are rearranged in a quincuncial shape to form a larger square.

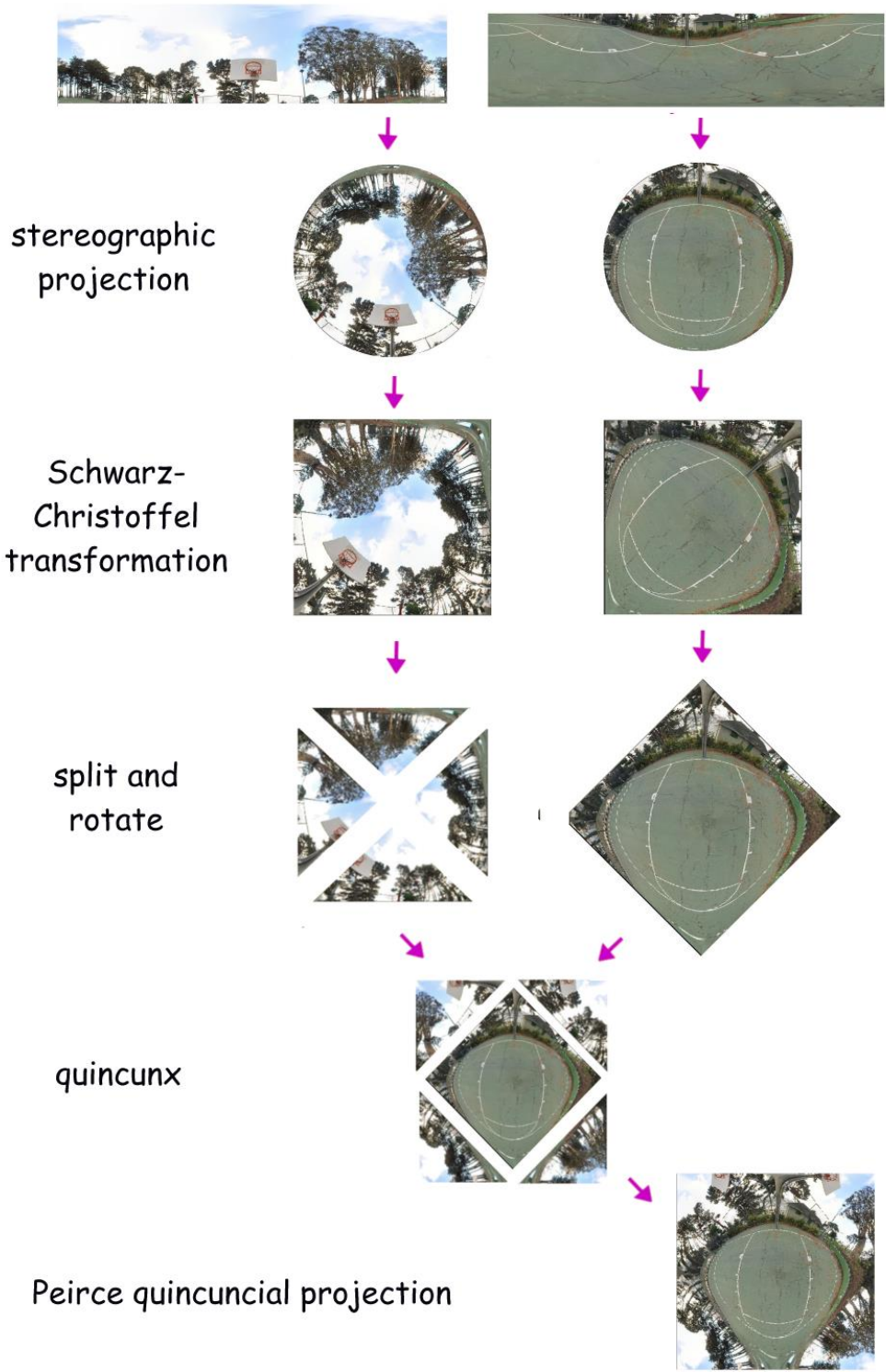

Figure 5: Visual example of the Peirce quincuncial projection steps

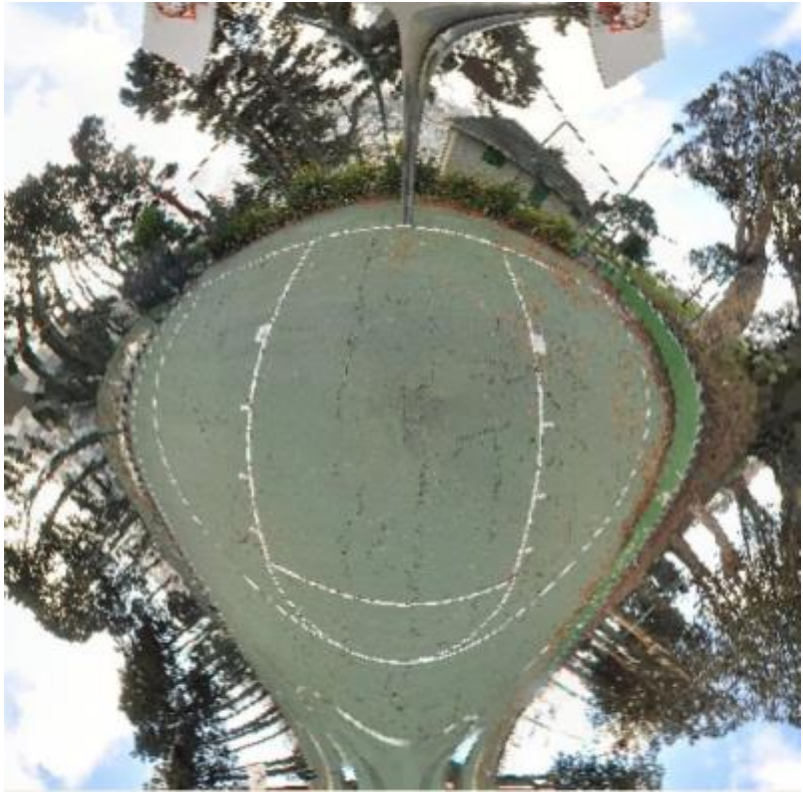

Figure 6: A Peirce quincuncial panorama

## 4 Limitations and Previous Work

As the panorama in figure 6 shows, the Peirce quincuncial projection has a serious inherent flaw. It is not conformal on four points. These four points fall on the midpoints in the perimeter of the square. For photographic applications, these non-conformal points cause ugly discontinuities in the image. This flaw has limited the use of the Peirce quincuncial projection in the spherical panorama community. For example, in figure 6, the non-conformal flaw causes the basketball ring to be split in half into left and right parts of the image. It also causes the bottom ground to bulge unnaturally.

During the authorship of this paper, there were only less than a dozen Peirce quincuncial panoramas showing up in a Google image search. In contrast, there were over 46,000 results for stereographic panoramas. This is true despite the fact that there is a Photoshop plugin called Flexify[9] by Flaming Pear Software for converting equirectangular images to Peirce quincuncial. Moreover, there is free source code available for a full implementation of the Peirce quincuncial projection in the Mathmap scripting plugin of the GIMP photo-editing application, written by flickr.com user Breic. It is likely that photographers who tried the Peirce quincuncial projection for their spherical panoramas found the non-conformal flaws unappealing and abandoned this projection. In this paper, we will present a proposed modification of the Peirce quincuncial projection to mitigate its non-conformal flaw. It is our hope that our modification will make the Peirce quincuncial just as popular as the stereographic projection among photographers.

Our proposed modification improves on previous ideas by people who have used the Peirce quincuncial projection for panoramas. For example, flickr.com member Breic[10] allows for the user to move the equator up and down in latitude within his Mathmap script. He also gives the user freedom to spin the spherical panorama along its main axis to move the four non-conformal points together around the equator before applying the projection. Our proposed modification gives the user more freedom by allowing the user to choose any latitude and longitude for each of the four non-conformal points. We will discuss a user-interface and an algorithm for warping the projection in the next section.

## 5 Warping the Equator

The main idea of this paper is to warp the equator of the input spherical panorama so that the non-conformal points fall on "uninteresting" areas of the image. This warping will hide the image discontinuities and make the result more aesthetically pleasing. The resulting image is a warped quincuncial panorama that is no longer conformal, but still generally preserves much of the characteristic look of an overhead aerial image.

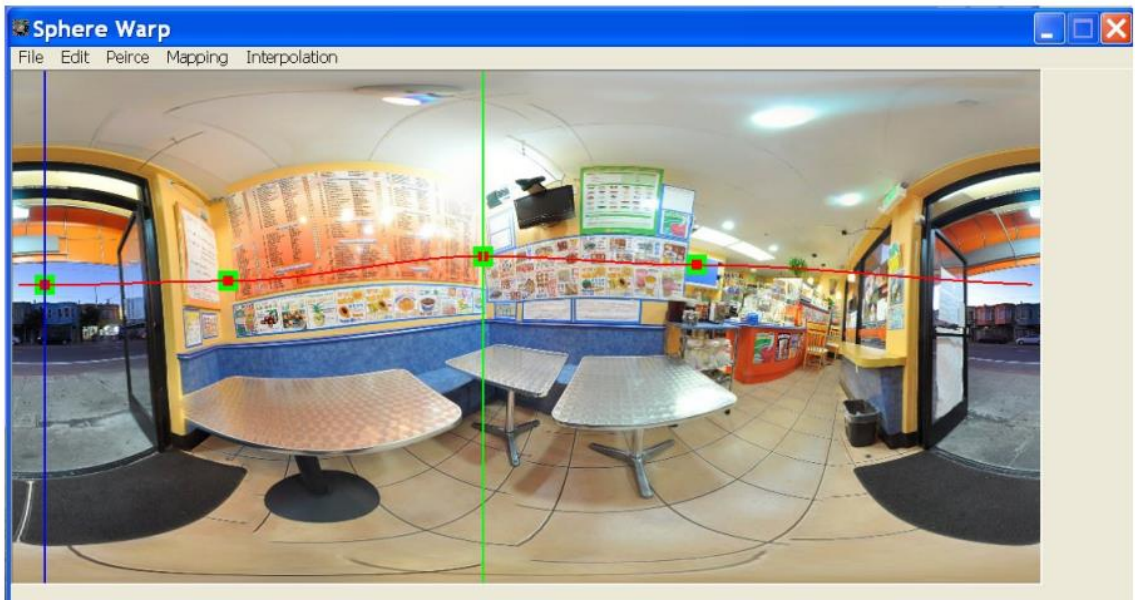

Figure 7: User-interface for selecting and moving four non-conformal points

We developed an interactive program where the user works on an input equirectangular image and moves four control points in the user-interface. These four control points always fall on the equator and will, in fact, redefine the equator after they have been selectively manipulated and moved. Moreover, these control points define the exact location of where the four non-conformal points of the projection will be mapped. Our program takes the location of these four control points and the input image to produce a warped Peirce quincuncial image. By careful selection of the control points, one can mitigate the discontinuities and distortions that arise in all Peirce quincuncial panoramas.

Figure 7 shows a screen shot of our program with its user-interface. The four control points define the warped equator (colored red). The first control point in the far left defines the international dateline (colored blue) which corresponds to a longitude of ±180° in the warped sphere. The third control point defines the prime meridian (colored green) which corresponds to a longitude of 0° in the warped sphere. The user is free to select and move these points to any pixel in the image.

After acquiring the coordinates of the four control points that define the warped equator, we can split the sphere into a warped northern hemisphere and a warped southern hemisphere. We can then proceed with the rest of the quincuncial pipeline which includes stereographic projections and Schwarz-Christoffel transformations.

It is important for the implementation of the Peirce quincuncial projection to be reasonably fast. We want to give the user quick feedback on the effects of manipulating control points to the resulting image. In practice, the bottleneck of the projection is the subroutine call to the complex-valued Jacobi elliptic function cn($z$,1/√2) for every output pixel. We will discuss ways to get around this computational bottleneck in a latter section of this paper and in the appendix.

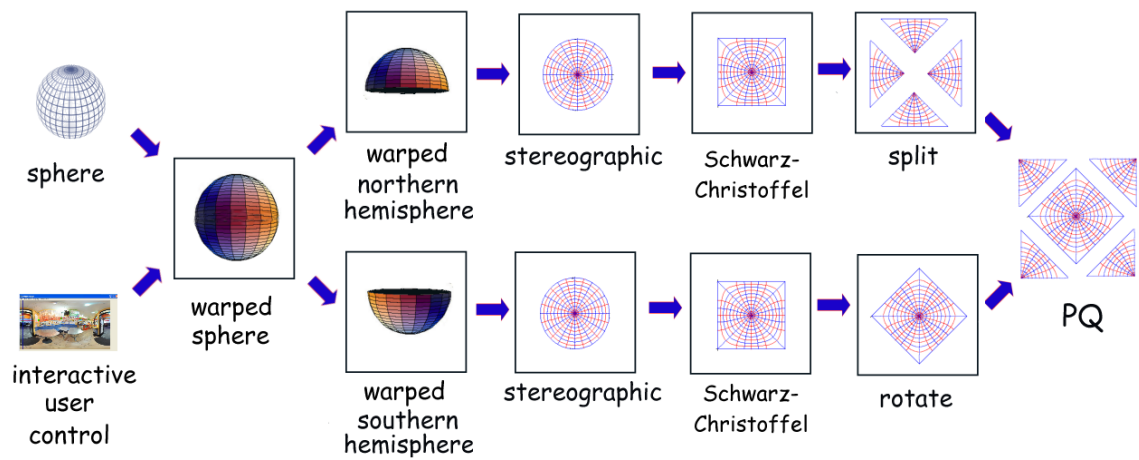

Figure 8: Pipeline for a warped Peirce quincuncial projection

Figure 8 shows an overview of our proposed pipeline for producing a warped Peirce quincuncial panorama. This pipeline adds an extra step to the one shown in figure 4 for computing the Peirce quincuncial projection. This extra step warps the input sphere under the guidance of the user through the careful selection of 4 control points. The rest of the downstream pipeline remains the same. The downstream pipeline includes the use of the stereographic projection and the use of the Jacobi elliptic function cn($z$,1/√2) for mapping the circular disc to a square.

## 6 Warping Algorithm

Before we talk about our warping algorithm, we need to discuss an important implementation detail. For efficiency reasons in imaging applications, it is actually desirable to calculate the Peirce quincuncial projection by reversing the steps given in the projection overview of figure 4. That is, we start from the output image, and reverse the mapping to fetch input pixels from the sphere to fill in the values in the output image. This is a lot more efficient than starting from an input spherical image and calculating where each input pixel get projected in the output image. Readers familiar with ray tracing or digital image resampling would observe that the same thing is done in those applications for the same efficiency reasons.

The main routine of a Peirce quincuncial implementation goes like this: For each pixel in the output image, calculate the latitude and longitude of the pixel in the sphere to fetch. It is in this calculation where one uses the equations of the stereographic projection and the Schwarz-Christoffel transformation. If the output pixel falls inside the central square of the quincunx, we fetch a pixel from the southern hemisphere. If the output pixel falls inside the four triangles on the corners of the quincunx, we fetch a pixel from the northern hemisphere.

Our warping implementation inserts a step before fetching a pixel value from the sphere. That is, given the latitude $\varphi$ and longitude $\lambda$ of the pixel to fetch in the sphere, we add a warping function $Warp(\varphi,\lambda)$ that produces warped $\varphi'$ and $\lambda'$ to replace our original latitude $\varphi$ and longitude $\lambda$ coordinates used for fetching pixels. The warping function $Warp(\varphi,\lambda)$ also has 4 user-provided parameters that control the warping process. We acquire these 4 parameters by providing a user-interface to manipulate four control points over an equirectangular image.

It is appropriate to mention here that for the sake of simplicity, we use linear interpolation in the description of our algorithm. Other types of interpolation schemes such as Catmull-Rom and Hermite spline interpolation with zero slope (m=0) also give satisfactory results, but in order to have simple equations, we just use linear interpolation in the algorithm explanation.

In linear interpolation, we usually have two input values *P* and *Q* which we want to interpolate, a control parameter $t \in [0.0,1.0]$, and an output value $R \in [P,Q]$. Without loss of generality, we assume *P<Q*. The main equation for linear interpolation involves a linear polynomial in *t*

$$R = (1 - t)P + tQ$$

For Catmull-Rom and Hermite spline interpolation, we use more complicated cubic polynomials in *t*, but the idea is the same in that the output $R \in [P,Q]$.

6.1 Four-points Warping Algorithm

We now define our warping function $Warp(\varphi, \lambda)$
   Input: latitude $\varphi$, longitude $\lambda$, four control points $(x_1,y_1)$, $(x_2,y_2)$, $(x_3,y_3)$, and $(x_4,y_4)$
   Output: warped latitude $\varphi'$, warped longitude $\lambda'$

Note that we label the control points as $(x_i,y_i)$ even though they actually correspond to points with longitudes and latitudes on the sphere in order to avoid confusion with our input and output latitude/longitude values.

1) Sort the input control points in longitude.

All our control points have longitude falling between -180° and 180°. In other words, we have $x_i \in$ [-180,180]. After sorting, each of $x_i$'s are arranged in increasing order. It is important for the user-interface to make each $x_i$ distinct. This means we should disallow the $x_i$'s to have the same longitude value. Mathematically speaking, the $x_i$'s should satisfy this inequality:

$$-180° \leq x_1 < x_2 < x_3 < x_4 \leq 180°$$

Since we are dealing with a spherical input image, we can actually define a phantom control point $x_5$ as $x_1 + 360°$. In the sphere, $x_1$ and $x_5$ fall on identical points.

2) Warp the longitude value

Warp longitude $\lambda$ to $\lambda'$ by first finding which quadrant $\lambda$ falls into. Knowing which quadrant $\lambda$ falls into allows you to select which control points to warp $\lambda$.

if $\lambda \in$ [-180,-90], select quadrant#1, assign lower bound L= -180 and upper bound U= -90
if $\lambda \in$ [ -90, 0],   select quadrant#2, assign lower bound L= -90  and upper bound U= 0
if $\lambda \in$ [0,90],      select quadrant#3, assign lower bound L= 0    and upper bound U= 90
if $\lambda \in$ [90,180],   select quadrant#4, assign lower bound L= 90   and upper bound U= 180

Given the quadrant number i, we have the control point longitude values as $x_i$ and $x_{i+1}$. Assign interpolation parameter t as

$$t = \frac{\lambda - L}{U - L}$$

Using linear interpolation over $x_i$ and $x_{i+1}$, we get the warped longitude as

$$\lambda' = (1-t)\ x_i + t\ x_{i+1}$$

3) Warp the equator latitude
At the start, the unwarped equator entirely falls on 0° latitude. However, we want to warp the equator latitude depending on the positions of the four control points. The warped equator will generally have a non-zero and varying latitude value that is dependent on longitude in the sphere. Let us denote the equator latitude as $\varphi_e$ which corresponds to the warped longitude $\lambda'$ that was calculated previously. Assign interpolation parameter t as

$$t = \frac{\lambda' - x_i}{x_{i+1} - x_i}$$

Using linear interpolation over $y_i$ and $y_{i+1}$, we get the warped equator latitude as

$$\phi_e = (1 - t)\, y_i + t\, y_{i+1}$$

4) Warp the latitude value
Once we have the warped equator value $\varphi_e$ from the previous step, we can warp our input latitude $\varphi$ to get $\varphi'$. There are two cases to consider depending on whether $\varphi$ is greater than or equal to zero (northern hemisphere) or less than zero (southern hemisphere).

case 4-1) southern hemisphere: $\varphi < 0°$

In the southern hemisphere, $\varphi \in [-90, 0]$. Assign interpolation parameter t as

$$t = -\frac{\phi}{90}$$

Using linear interpolation, we get the warped latitude as

$$\phi' = (1 - t)\, \phi_e - 90\, t$$

case 4-2) northern hemisphere: $\varphi \geq 0°$
In the northern hemisphere, $\varphi \in [0, 90]$. Assign interpolation parameter t as

$$t = \frac{\phi}{90}$$

Using linear interpolation, we get the warped latitude as

$$\phi' = (1 - t)\, \phi_e + 90\, t$$

5) Output the warped latitude $\varphi'$ and warped longitude $\lambda'$

Our warping algorithm works with angles in degrees instead of radians. We decided to use degrees for easier debugging. The implementation can easily be converted to work in radians if so desired by changing hardcoded constants of -180°, -90°, 90°, and 180° to corresponding values in radians.

We refer to our warping algorithm as the *4-points warping algorithm* because it is based on warping the sphere through four control points on the sphere. This warping algorithm is independent of the internals of the Peirce quincuncial projection and can easily be applied to other map projections as well. We have, however, tailored the warping for mitigating non-conformal discontinuity artifacts in the Peirce quincuncial projection. One can think of this warping step as a preprocessing step before applying the Peirce quincuncial projection.

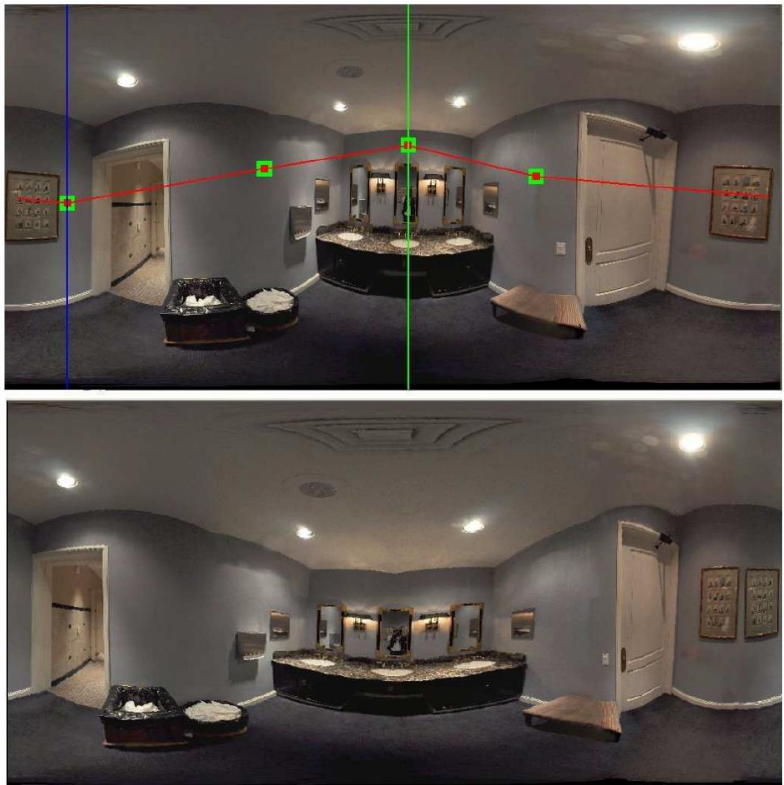

Figure 9: An equirectangular panorama (top) and its warped version (bottom)

## 6.2 Slerp: Spherical Linear Interpolation

The reader might ask why our warping algorithm uses linear interpolation or Catmull-Rom interpolation on the sphere when there is a much better alternative known to many: spherical linear interpolation, also known as slerp.

This is a fair question which we shall address here. The simple answer is, we do not need to. Our warping algorithm takes in inputs and produces outputs in spherical coordinates of latitude and longitude. Slerp is only valid for interpolating Cartesian coordinates. Since we never depart from spherical coordinates in our algorithm, we do need to use slerp.

# 7 Results

We now provide visual comparisons of our Peirce quincuncial projection resulting from an unwarped and warped equator. Figure 10 shows our results applied to the input image from figure 7.

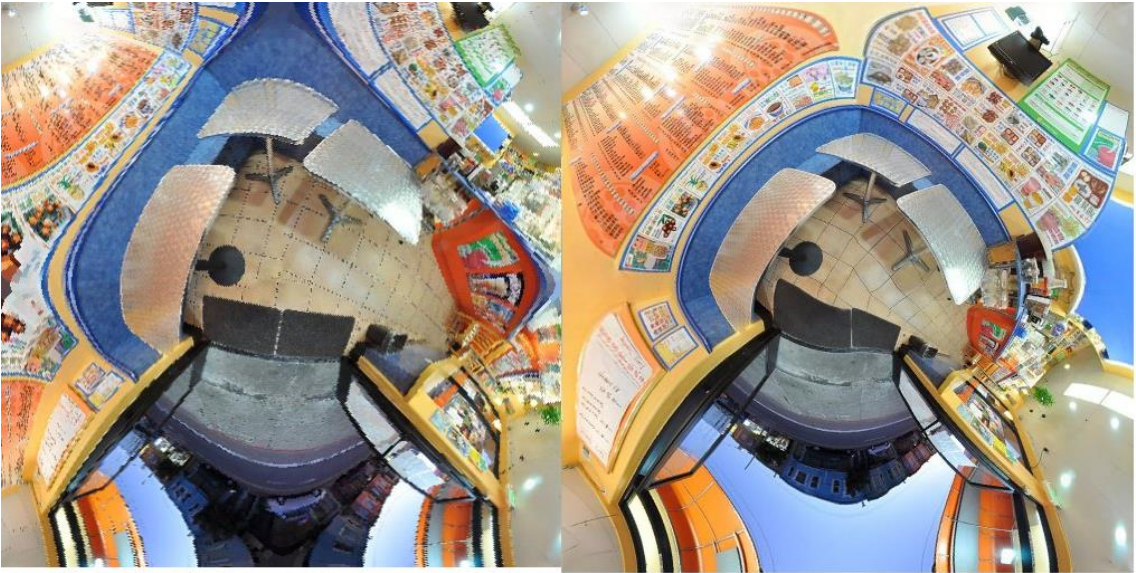

Figure 10: A visual comparison of an unwarped and warped Peirce quincuncial projection

We provide more results in Figures 11a, 11b, and 11c. These figures contain an equirectangular input image on top with four control points carefully positioned; followed by an unwarped and warped Peirce quincuncial projection at the bottom.

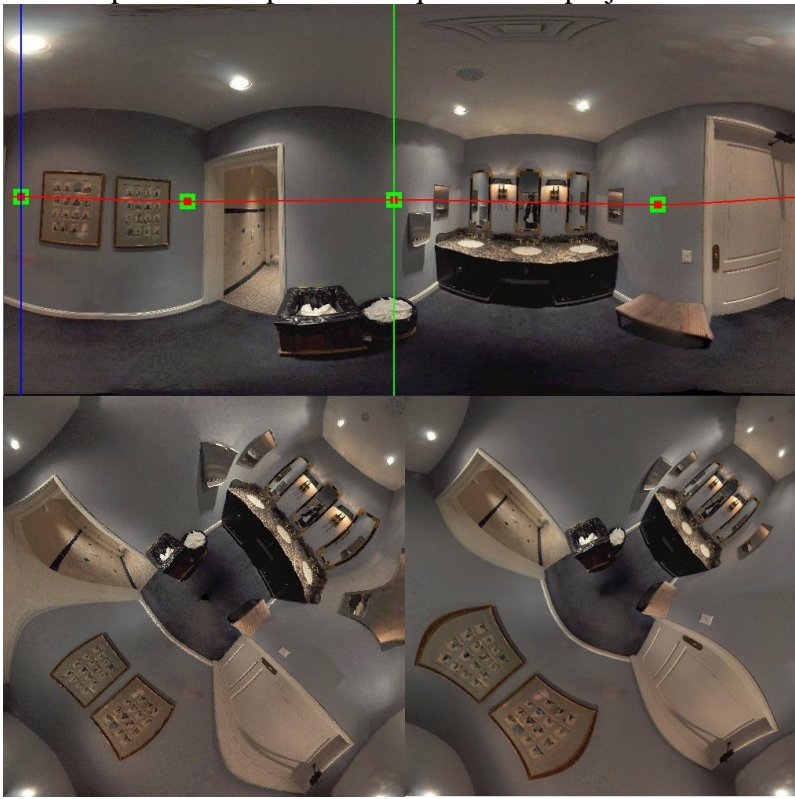

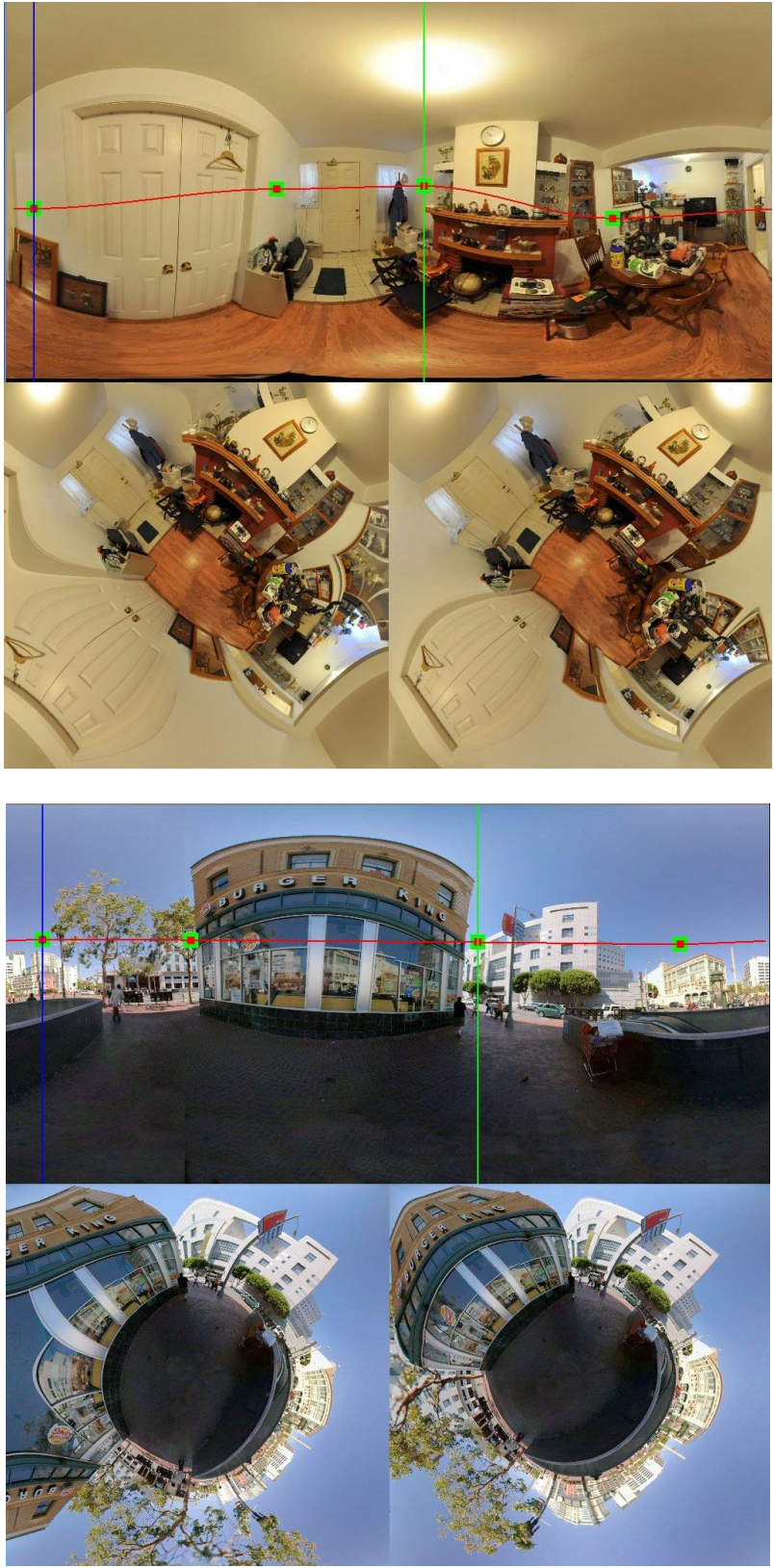

Figure 11abc: More comparisons of unwarped and warped Peirce quincuncial panoramas

# 8 A Fast Peirce Quincuncial Algorithm for Images

In order to facilitate an interactive user-interface for warping a Peirce quincuncial panorama, we need to perform the projection sufficiently fast. Each time the user moves a control point, we want the computer to be able to recompute the projection and give quick user-feedback. In this section, we present a fast algorithm for computing the Peirce quincuncial projection of spherical images.

The simplest way to speed up the computation of the Peirce quincuncial projection is by precomputing the mapping in uniformly sampled points in the complex plane. Using this approach, one can build a lookup table which gives latitude and longitude values for every pixel in the output image. This was the approach done by German[3] et.al. in their implementation of the Peirce quincuncial projection. As a matter of fact, Peirce[1] also resorted to a precomputed lookup table in his original derivation of the quincuncial projection. Afterwards, interpolation is used to fill in the blanks for in-between values not provided in the precomputed table.

We have found that using a precomputed table with interpolation gives unsatisfactory results especially when the input spherical image is warped. We almost always have to warp the image in order to mask the ugly non-conformal points of the projection. We, therefore, need a reasonably fast algorithm for computing the Peirce quincuncial projection without resorting to precomputed values stored in a table.

## 8.1 Dihedral Symmetry of the Square

The square has 8 different symmetries that correspond to the $D_8$ group. The symmetries arise from rotations by 90°; and from reflections along the main horizontal, vertical, and diagonal axis. Figure 12 shows the square and its symmetries. It is possible to use these symmetries to cut down on the computations needed in performing a Peirce quincuncial projection. Instead of performing calculations on the whole square, we can use symmetry and just perform calculations on a semi-quadrant of the square

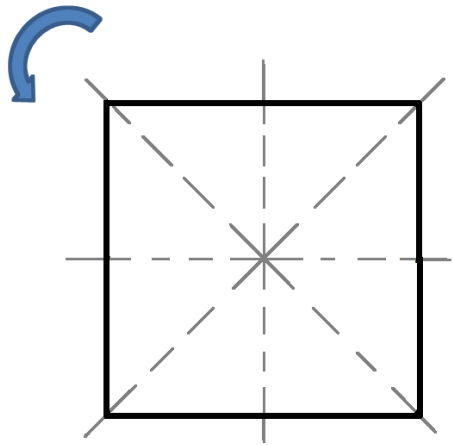

Figure 12: Dihedral symmetry of the square

## 8.2 Sixteenfold Symmetry of the Peirce Quincunx

Recall that the Peirce quincunx in Figure 4 is formed by joining two squares. One square is tilted diamond inside the quincunx and the other is split as the petals of the quincunx. Each square corresponds to northern or southern hemispheres of the full sphere. Since each square has eightfold symmetry and these squares are north-south mirror images of each other, we essentially have sixteenfold symmetry in the Peirce quincunx. It is possible to use this symmetry to avoid the evaluation of a complex Jacobi elliptic function $cn(z,1/\sqrt{2})$ for every output pixel in our projection. Instead, we only do the required calculations in a basic aperiodic triangular region of the Peirce quincunx. This triangular region corresponds to a semi-quadrant of a hemispherical square. We shall denote this region as the *fundamental region* of the Peirce quincunx. After evaluating a complex Jacobi elliptic function on this fundamental region, we can use rotation and reflection symmetries to fill in the rest of the Peirce quincunx. Figure 13 shows the fundamental region in relation with the whole Peirce quincunx.

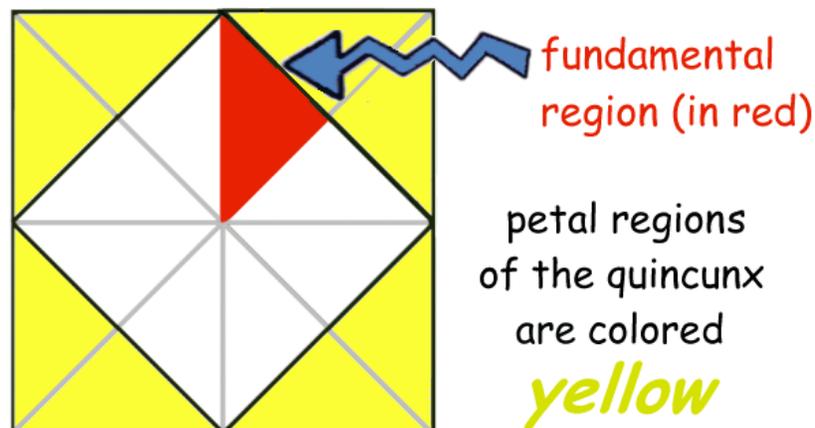

Figure 13: fundamental region of the Peirce quincunx

## 8.3 Pseudo-Code

Here is pseudo-code describing an outline of our fast algorithm for computing the Peirce quincuncial projection. For more details, see the appendix for actual Matlab code that implements the fast Peirce quincuncial projection on an input grayscale image.

for each pixel in the fundamental region
- (1) Calculate a complex Jacobi elliptic function $cn(z,1/\sqrt{2})$ and inverse stereographic projection to map pixel coordinates to latitude/longitude coordinates on the surface of the sphere
- (2) Use reflection to fill the other half of the semiquadrant in quadrant 1. This can be done by simply negating the already calculated longitude from step (1)
- (3) Use 90° rotation of values from quadrant 1 to fill in quadrant 2
- (4) Use 180° rotation of values from quadrant 1 to fill in quadrant 3
- (5) Use 270° rotation of values from quadrant 1 to fill in quadrant 4
- (6) Use reflection to fill in the other hemispherical square in the petal regions. This can be done by simply negating the already calculated latitude from step (1)

end

## 9 Motivation

We advocate for the use of Peirce quincuncial projection as a viable alternative to the stereographic projection for spherical panoramas. In this section, we justify this proposal. We give four reasons for it.

### 1) Full Mapping of the Sphere

The stereographic projection cannot map the entire sphere into a finite image. This is because one pole maps to infinity in the stereographic projection. In general, the coverage of a stereographic projection is $4\pi - \varepsilon$ steradians. Epsilon $\varepsilon$ is defined as a small region around one pole that has to be cropped in order to get a finite image. For outdoor images, this cropping is usually not noticeable and can be ignored. This is because the cropped region in outdoor images is usually just a portion of the homogeneous blue sky. In contrast, for indoor images, the cropped region is usually a part of the ceiling. Cropping a portion of the ceiling out sometimes makes the indoor spherical panorama appear incomplete.

Moreover, since the Peirce quincuncial projection can map all $4\pi$ steradians of the sphere, this means that it is possible to do an inverse projection from a Peirce quincuncial image and get the original spherical panorama back. In stereographic panoramas, this is impossible because of the missing cropped region.

### 2) No Excessive Enlargement of Objects near the Opposite Pole

For indoor and outdoor panoramas, tall structures will undergo excessive enlargement in stereographic images. Likewise, features high up in the ceiling or in the sky will appear very large if not cropped from the stereographic image. This is a common problem that manifests itself in stereographic panoramas with buildings and tall shelves. Figure 10 shows examples of stereographic images with this problem. In these images, the enlarged features dwarf the rest of the objects in the panorama. This disproportionate enlargement usually makes the stereographic panorama visually unappealing. Figure 11 shows a comparison of a stereographic panorama with ceiling enlargement and its corresponding Peirce quincuncial panorama. Later, we will devote a whole section to discuss a possible solution to the stereographic enlargement problem via a hybrid projection.

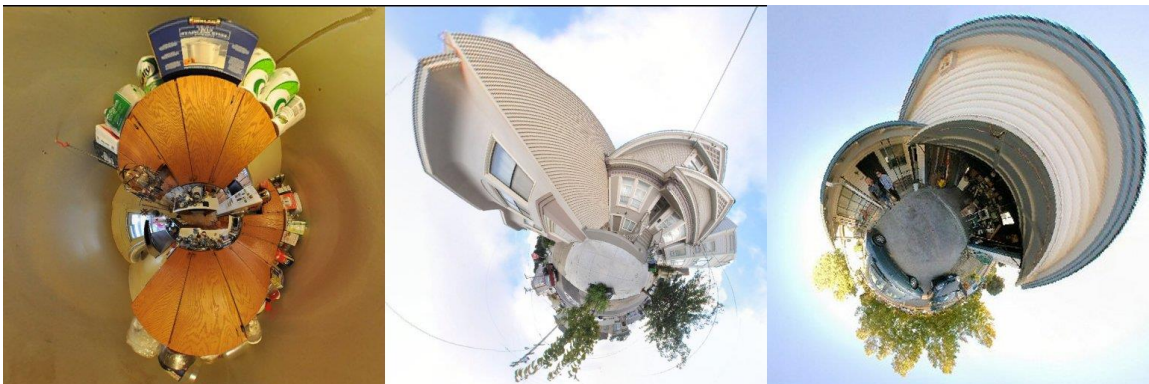

Figure 14: Stereographic enlargement of a kitchen cupboard and two houses

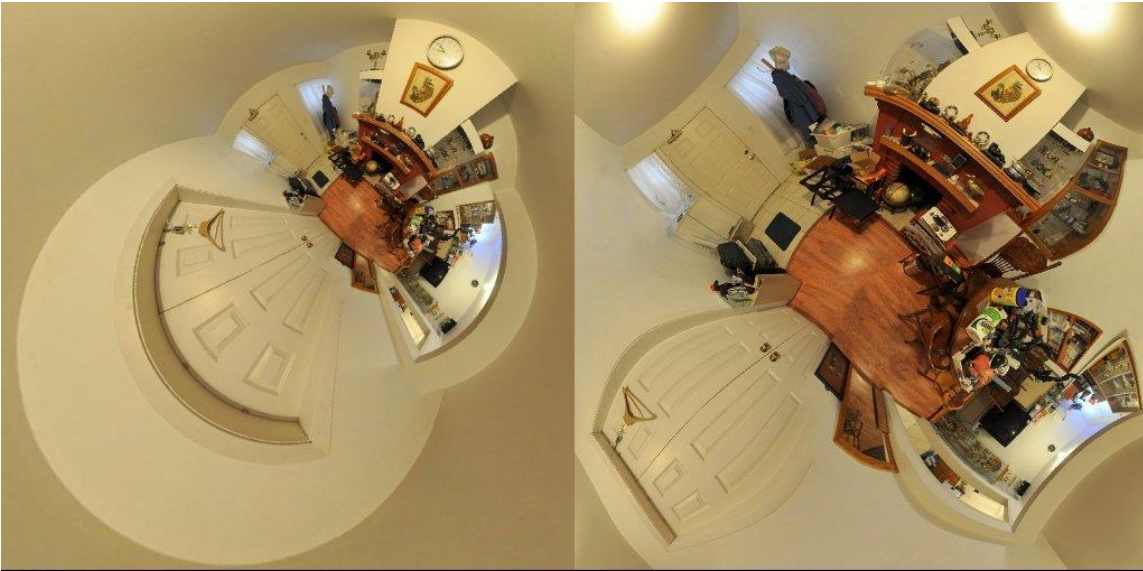
Figure 15: less pronounced ceiling enlargement in the Peirce quincuncial projection (left)

    3) Four Corners in a Room

Most indoor panoramas are inside a room with four walls and four corners. The Peirce quincuncial projection is natural to represent these panoramas because the square has four corners. In fact, the four corners of the room usually map directly to the four corners of the square. Figures 15 and 16 are examples of indoor panoramas where the Peirce quincuncial projection is more appropriate than the stereographic projection. Stereographic projections tend to be circular and cause rooms to lose their natural rough edges along the four corners.

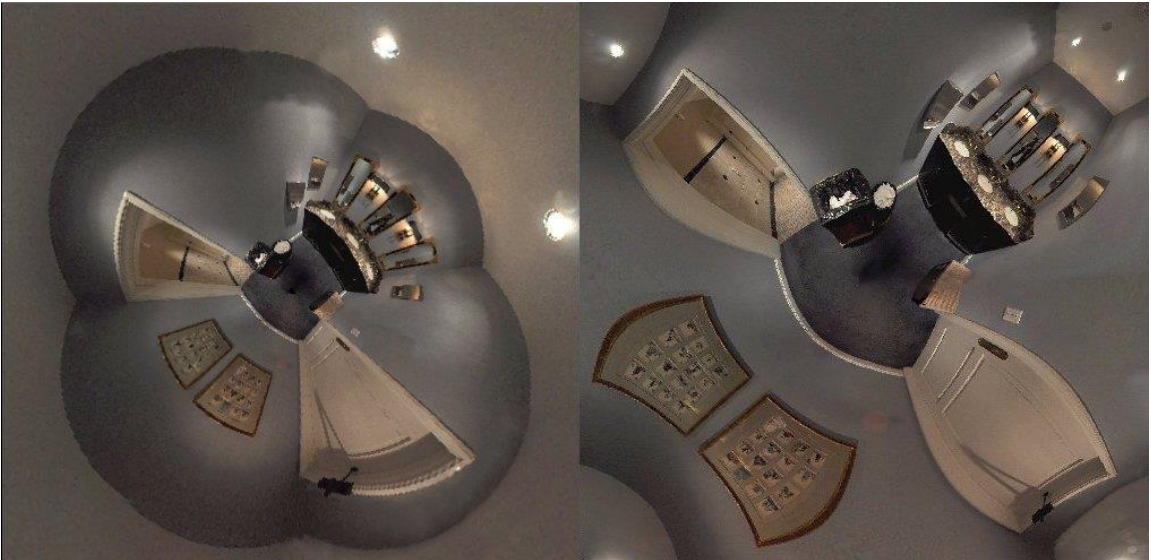
Figure 16: A comparison of the stereographic with the Peirce quincuncial projection

#### 4) Tileable Panoramas

One important property of the Peirce quincuncial projection is its tileability. One can arrange Peirce quincuncial panoramas to tile the plane regularly like a wallpaper. In fact, flickr.com user Breic[10] implemented his Mathmap script just for this reason. Many of the Peirce quincuncial images in the internet are abstract photographs consisting of tiled Peirce quincuncial panoramas.

## 10 Discussion:

We originally tried to come up with an algorithm to determine the placement of the four non-conformal points automatically. However, we were unable to match the results we were getting when manually moving the non-conformal points. Ultimately, we decided to relegate the task of automating the placement of the four non-conformal points as future work. We did come up with some heuristics for this task. These heuristics can be used as a guideline for artists in warping Peirce quincuncial panoramas or by algorithm designers who want to take on the problem of automating the task.

Heuristics:

1) Select a location where there is very little illumination and color variance in the image. This means that one can use edge detection and image segmentation algorithms to mark edge boundaries that should be avoided. For indoor panoramas, one can select a homogeneous wall as a non-conformal location. For outdoor panoramas, the sky works well as a non-conformal location.

2) We have an exception to heuristic#1. Most indoor panoramas are inside a room with four walls and four corners. Sometimes, positioning the non-conformal points directly on the corner edges of the room works well.

3) The eye is more forgiving to image discontinuities if the resulting warped panorama has symmetry along these discontinuities. Recall that the four non-conformal points fall along the edge midpoints on the perimeter of the Peirce square. Preserving left/right and top/down symmetry along these edges makes the non-conformal discontinuities more palatable to the eye.

4) The entire equator should have a latitude value close to 0°. If one moves the equator too high up or too low, the resulting Peirce quincuncial panorama will tend to look like a stereographic projection. Of course, this is not necessarily bad. In fact, we will talk about this more in the next section. Nevertheless, we want to emphasize that putting the equator near the horizontal middle gives the Peirce quincuncial projection its distinct look.

5) The equator should remain horizontal as much as possible. Excessive bending of the equator causes more pronounced warping artifacts in the panorama.

6) The four control points should stay equidistant with their neighboring control points for as much as possible. Positioning control points too close together in longitude tends to cause more pronounced area distortions in the warped panorama.

## 11 The Antipode Perimeter Square Projection

We would like devote a whole section to discuss a special case of the warped Peirce quincuncial projection. When one moves the four control points too high up or too low in latitude, the resulting Peirce quincuncial image resembles a stereographic projection more and more. In the limiting case where all four control points are at 90° latitude, the resulting warped Peirce quincuncial image will map the entire sphere to the central square of the quincunx. The four triangles on the corners of the quincunx no longer map to a hemisphere. In fact, we can discard these four triangles and just use the oblique central square of the quincunx. Moreover, if we spin the four control points in longitude by 45°, we can get an aligned unrotated square. We call this special case of the warped Peirce quincuncial projection as the *Antipode Perimeter Square (APS) projection*. It comes in two different flavors depending on whether the four control points are at +90° latitude (zenith case) or at -90° latitude (nadir case). We shall limit our discussion here to the zenith case. The left image in figure 17 shows a warped Peirce quincuncial projection when all four control points are at +90° latitude applied to the input image in figure 1. The right image in figure 17 shows the Antipode Perimeter Square projection where all four corner triangles of the quincunx are truncated and the control points are adjusted to align the square.

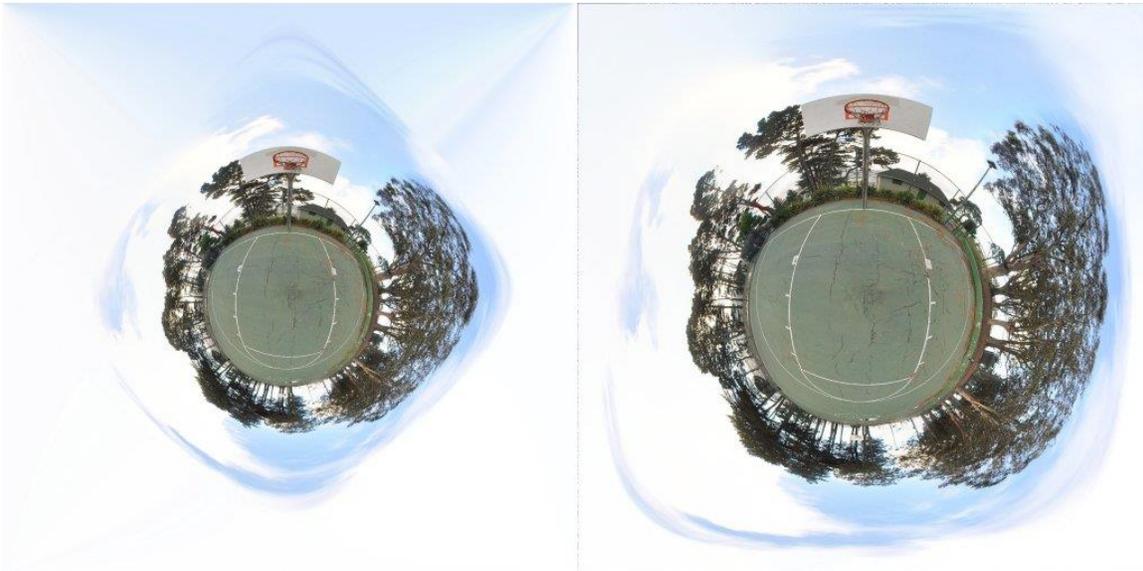

Figure 17: Warped Peirce quincuncial panorama with all control points at 90° latitude (left) and an Antipode Perimeter Square panorama (right)

Let us discuss some properties of the APS projection. The APS projection is a hybrid between the Peirce quincuncial projection and the stereographic projection. In fact, it resembles the stereographic projection more but has some nice properties that it inherited from a warped Peirce quincuncial projection. For one thing, it is a way to get a stereographic projection of the whole sphere to fit into a finite image. Recall that the stereographic projection maps one pole to infinity and cannot represent the whole sphere in a finite planar image. The APS projection, on the other hand, maps one pole to the whole perimeter of a square. Moreover, the perimeter of the square is the only part of this projection that has non-conformal discontinuity. In effect, the APS projection uses the Schwarz-Christoffel transformation to convert a warped sphere into a square.

The APS projection does not piece together two rectified hemispheres in a quincuncial manner and does not have the four non-conformal points of the Peirce quincuncial projection. Moreover, the APS projection requires less computation because only one square mapping is calculated via the Schwarz-Christoffel transformation.

The APS projection is a viable solution to the problem of excessive enlargement of objects in the stereographic projection. In figures 18, 19, 20, and 21, we show visual comparisons of the stereographic projection (left) with the APS projection (right) for indoor and outdoor scenes.

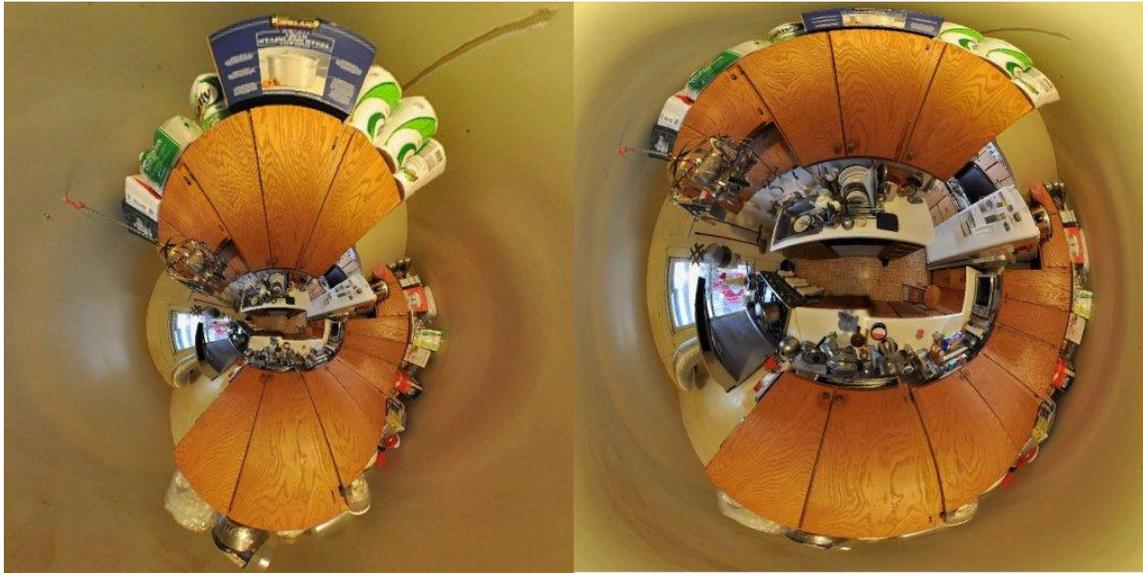

Figure 18: The APS panorama (right) exhibits less enlargement of objects near the ceiling. Hence, features near the ground have more detail in the panorama.

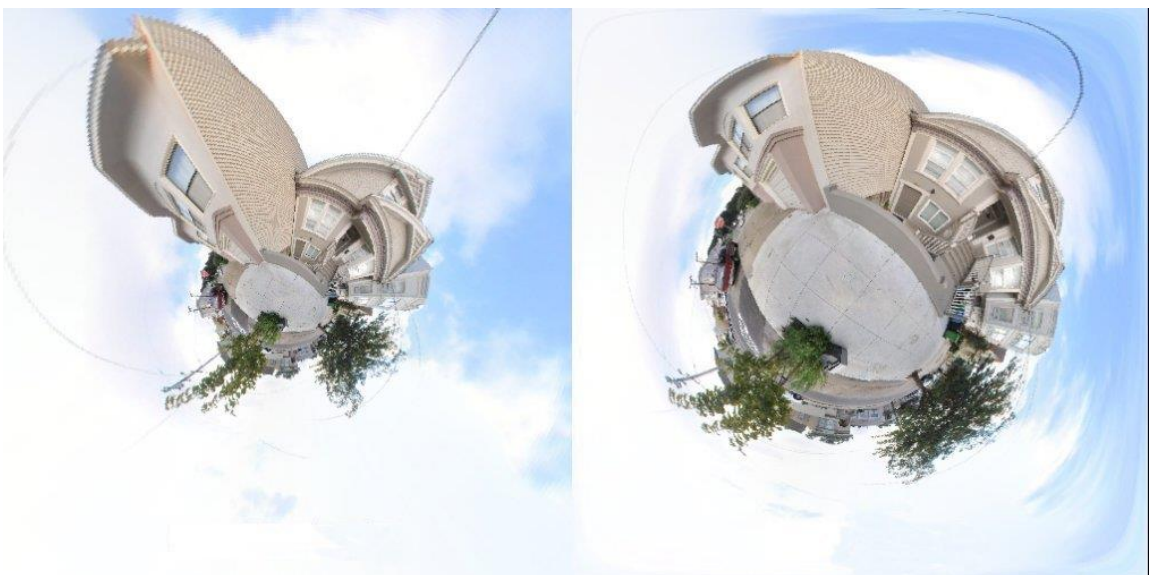

Figure 19: The APS panorama (right) exhibits less enlargement of the tall house.

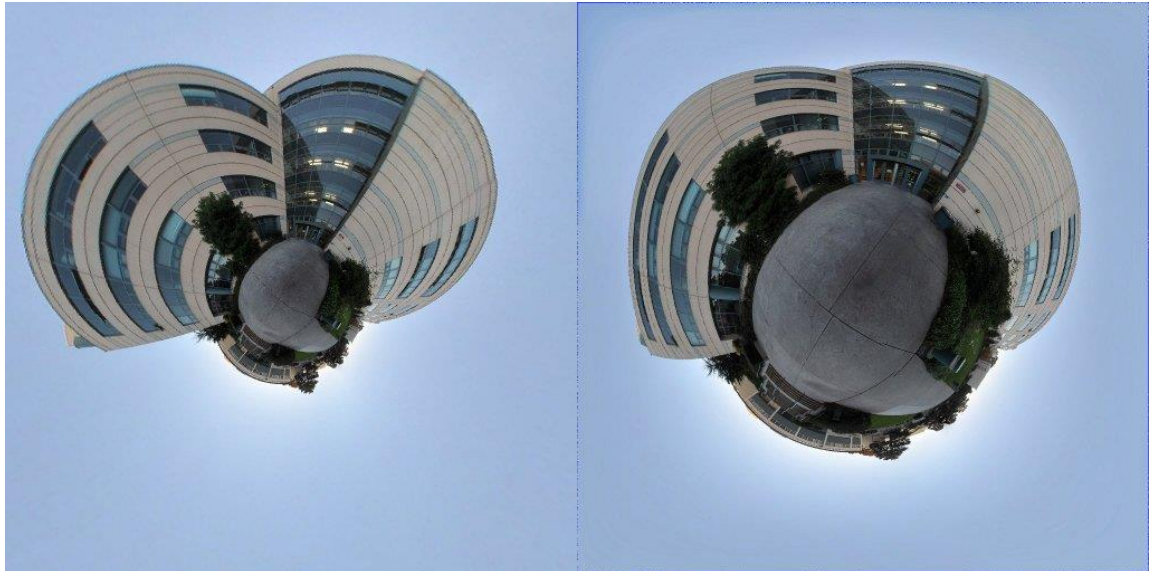

Figure 20: Excessive fanning enlargement of the stereographic buildings (left)

In the left image of figure 20, we see a common problem of stereographic panoramas with tall buildings and skyscrapers. There is excessive enlargement of the building near the top and the building shape tends to look like a fan. This problem can be alleviated by moving the opposite pole of the projection closer to the building. Normally, this opposite pole of the projection is at the nadir of the sphere and at the center of the projected image. Some stereographic software, such as Hugin, allow the user to move this point. Moving this point does reduce the excessive enlargement and fanning of the buildings. However, this solution only works if one is not surrounded by tall buildings in other directions. We believe that the APS projection is a better solution to the excessive enlargement and fanning problem.

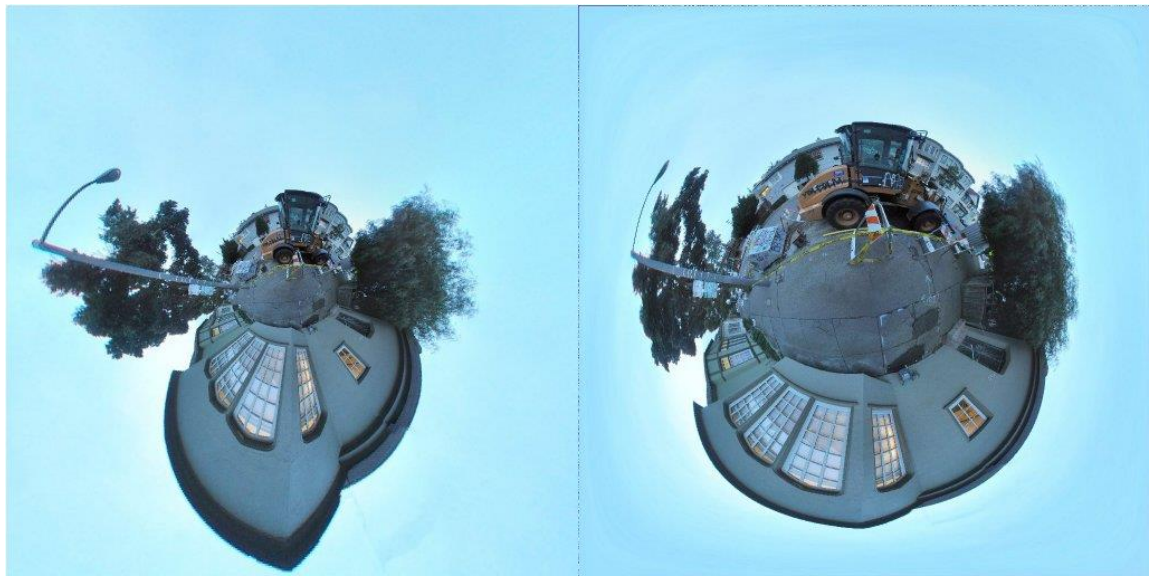

Figure 21: The APS projection (right) has a better balance of detail between the top and bottom hemispheres of a spherical panorama

The APS projection distorts area less than the stereographic projection. This results in a better balance of detail between features in the northern and southern hemispheres of the projected spherical panorama. Recall that the stereographic projection maps $4\pi - \varepsilon$ steradians of the sphere onto the plane. $\varepsilon$ (epsilon) can be chosen to be as small as one wishes, at the expense of area distortion. In general, the smaller $\varepsilon$ is, the more disparity in size there will be between the projected northern hemisphere and the projected southern hemisphere.

We believe that the spherical panorama community should consider using the APS projection when creating their "little planets". This could result in a better rendering of their art. In fact, we can imagine that future versions of spherical panorama tools such as Hugin and Panotools will have the APS projection as a feature. This will enhance the tool chest and repertoire of photographers everywhere.

## 12 Future Work

As previously mentioned, it would be ideal to have an algorithm for automated placement of the four non-conformal points of the projection. We can use techniques from computer vision and machine learning to accomplish this.

The Guyou projection and the Adams hemisphere-in-a-square projection are conformal projections related to the Peirce quincuncial projection in the use of the Schwarz-Christoffel transformation for spherical mapping. Just like the Peirce quincuncial projection, these projections have four points of non-conformality that can cause unsightly imaging artifacts. Our warping algorithm can be modified and extended for both these projections.

In computer graphics, spherical panoramas are frequently used in environment mapping[11] and high dynamic range imaging. We can foresee the use of the Peirce quincuncial projection for such applications. Moreover, since the Peirce quincuncial projection has a square shape, it can easily be adapted to use with mipmaps and clipmaps.

## 13 Conclusion

We presented a modification to the Peirce quincuncial projection for photographic applications. This modification allows the user to position four non-conformal points on the sphere through an interactive user-interface. We then presented an algorithm to warp the Peirce quincuncial image based on these four control points. We argued that our proposed modification makes the Peirce quincuncial projection an appealing alternative to the stereographic projection for indoor panoramas and some outdoor panoramas. We then presented a fast algorithm for computing the Peirce quincuncial projection. We also presented and examined a special case of the warped Peirce quincuncial projection called the Antipode Perimeter Square (APS) projection.

For more examples of Peirce quincuncial panoramas, visit the Peirce quincuncial group in flickr: http://www.flickr.com/groups/quincuncial/

# Appendix A: Matlab code

In this appendix, we provide Matlab source code for calculating the Peirce quincuncial projection of a spherical image. There are 3 files in this package

1) slowpquin.m – a script for the Peirce quincuncial projection without our optimization
2) insitupq.m – a script for converting (x,y) square coordinates to $(\phi, \lambda)$ sphere coordinates
3) fastpquin.m – an implementation of our fast Peirce quincuncial algorithm

Note that the files slowpquin.m and fastpquin.m rely on insitupq.m as a subroutine. Meanwhile, slowpquin.m and fastpquin.m are independent of each other. One can use one or the other to compute the Peirce quincuncial projection of a spherical image. The latter is up to 16 times faster because of its use of dihedral symmetries.

Both slowpquin.m and fastpquin.m take as input a gray-scale equirectangular image in the form of a Matlab matrix. The aspect ratio of the image must be 2x1. The output is a square image in the form of a matrix of gray-scale values.

For those, who intend to try these scripts in Matlab. We recommend the use of these other Matlab commands for conjunction with our scripts
1) imread – for reading in images
2) imagesc – for displaying images
3) colormap – for lookup table of gray images

```matlab
function v=slowpquin(sphimage)
% slow Peirce quincuncial projection of spherical imagery
% input: grayscale equirectangular image
% output: Peirce quincuncial projection
%
rows=size(sphimage,1);
columns=size(sphimage,2);
range=1;
xdomainmin = -range;
xdomainmax =  range;
ydomainmin = -range;
ydomainmax =  range;
outrows = rows;
outcolumns = outrows;
v=zeros(outrows,outcolumns);
for yo=1:outrows
  for xo=1:outcolumns
    xi = ((xdomainmax-xdomainmin)/(outcolumns-1)*(xo-1)+xdomainmin);
    yi =  ((ydomainmin-ydomainmax)/(outrows-1)*(yo-1)+ydomainmax);
    [latp,longd] = insitupq(xi,yi);
    longd = mod(longd+180,360)-180;
    v(yo,xo)=indexImage(sphimage,latp,longd);
  end
end

function value = indexImage(sphimage,latp,longd)
   rows=size(sphimage,1);
   columns=size(sphimage,2);
   xindex=round((longd+180)*(columns-1)/360+1);
   yindex=round((latp-90)*-(rows-1)/180+1);
   if longd <= 180 & longd >= -180 & latp <= 90 & latp >= -90
      value=sphimage(yindex,xindex);
   else
      value = 128;
   end
return
```

```matlab
function [latp,longd] = insitupq(x,y)
% in situ Peirce Quincuncial inverse projection
% input: x and y coordinates inside Peirce square [-1,1] x [-1,1]
% output: latitude and longitude values on the sphere
%
% This script uses the Schwarz-Christoffel transformation to map
%   a unit disc to a square conformally
% There are many other ways to map a disc to a square.
% See the paper "Analytical Methods for Squaring the Disc" for
%   a detailed discussion of other mappings
%   ( including the Schwarz-Christoffel mapping )
% http://arxiv.org/abs/1509.06344
%
% Also see more example pictures in
%   http://peircequincuncial.blogspot.com/
%   http://squircular.blogspot.com
%
% ke is complete elliptic integral of the 1st kind at m=1/2
ke = 1.85407467730137;
m = 1/2;
tolerance = 0.0001;

% call to complex-valued Jacobi elliptic cn function
% see Abramowitz-Stegun (AMS55) for the equations
if (abs(y)<tolerance)
    [sni,cni,dni] = ellipj(ke * (x - 1), m, tolerance);
    rex = cni;
    imy = 0.0;
else
    phi = ke * (x - 1);
    psi = ke * y;
    [s,c,d] = ellipj(phi, m, tolerance);
    [s1,c1,d1] = ellipj(psi, 1-m, tolerance);
    delta = c1^2 + m * s^2 * s1^2;
    rex =(c * c1) / delta;
    imy = -(s * d * s1 * d1) / delta;
end

% rex and imy are coordinates in the disc
% which can then be converted to spherical geodetic coordinates

 % inverse stereographic projection
latp = ( 2 * atan(sqrt(rex^2 + imy^2)) - pi/2);
longd = atan2(rex,imy);
latp = latp * 180/pi;
longd = longd * 180/pi;
```

```
function v=fastpquin(sphimage)
% fast Peirce quincuncial projection using sixteenfold symmetry
% input: grayscale equirectangular image
%        aspect ratio should be 2:1
%        width should be divisible by 4;
%        height should be an even number
% output: Peirce quincuncial projection of the input
%
rows=size(sphimage,1);
columns=size(sphimage,2);
range=1;
xdomainmin = -range;
xdomainmax =  range;
ydomainmin = -range;
ydomainmax =  range;
outrows = rows;
outcolumns = outrows;
halfrows = outrows /2.0;
halfcolumns = outcolumns / 2.0;
columns3o4 = outcolumns * 3.0/4.0;
v=zeros(outrows,outcolumns);
ylow = 0;
yhigh = halfrows+2;
for xo=halfcolumns:columns3o4
  ylow = ylow + 1;
  yhigh = yhigh -1;

  for yo= ylow:yhigh
    xi = ((xdomainmax-xdomainmin)/(outcolumns-1)*(xo-1)+xdomainmin);
    yi =  ((ydomainmin-ydomainmax)/(outrows-1)*(yo-1)+ydomainmax);
    [latp,longd] = insitupq(xi,yi);
    longdr = mod(longd+180,360)-180;
    longdl = mod(-longd+180,360)-180;

%%%%%%%%%%%%%%%%%%%%%%%%%%%%%%%%%%%%%%%%%%%%%%%
% top triangle
    %%   inner diamond
    v(yo,xo)=indexsphere(sphimage,latp,longdr);
    v(yo,outcolumns-xo+1)=indexsphere(sphimage,latp,longdl);

    %%% outer petal region upper
    v(xo-halfcolumns+1,yo+halfcolumns) = indexsphere(sphimage,-latp,longdr);
    v(xo-halfcolumns+1,halfcolumns-yo+2) = indexsphere(sphimage,-latp,longdl);

%%%%%%%%%%%%%%%%%%%%%%%%%%%%%%%%%%%%%%%%%%%%%%%
% left side triangle
    longdla = -90+longdr;
    longdlb = -90+longdl;
    longdlr = mod(longdla+180,360)-180;
    longdll = mod(longdlb+180,360)-180;

    % upper
    v(outcolumns-xo+1,yo) = indexsphere(sphimage,latp,longdlr);;
    % lower
    v(xo,yo) = indexsphere(sphimage,latp,longdll);
```

```
      % outer petal region left side
      v(halfcolumns-yo+2,xo - halfcolumns+1) = indexsphere(sphimage,-latp,longdlr);
      v(halfcolumns+yo,xo - halfcolumns+1) = indexsphere(sphimage,-latp,longdll);

%%%%%%%%%%%%%%%%%%%%%%%%%%%%%%%%%%%%%%%%%%%
% right side triangle
      longdra = 90+longdr;
      longdrb = 90+longdl;
      longdrr = mod(longdra+180,360)-180;
      longdrl = mod(longdrb+180,360)-180;

      % upper
      v(outcolumns-xo+1,outcolumns-yo+1) = indexsphere(sphimage,latp,longdrl);;
      % lower
      v(xo,outcolumns-yo+1) = indexsphere(sphimage,latp,longdrr);

      % outer petal region right side
      v(halfcolumns-yo+2,outcolumns+halfcolumns-xo+1) = indexsphere(sphimage,-latp,longdrl);
      v(halfcolumns+yo,outcolumns+halfcolumns-xo) = indexsphere(sphimage,-latp,longdrr);

%%%%%%%%%%%%%%%%%%%%%%%%%%%%%%%%%%%%%%%%%%%
% bottom triangle
      longdba = 180+longdr;
      longdbb = 180+longdl;
      longdbr = mod(longdba+180,360)-180;
      longdbl = mod(longdbb+180,360)-180;

      % right
      v(outcolumns-yo+1,xo)= indexsphere(sphimage,latp,longdbl);
      % left
      v(outcolumns-yo+1,outcolumns-xo+1)= indexsphere(sphimage,latp,longdbr);

      % outer petal region below
      v(outcolumns+halfcolumns-xo+1,yo+halfcolumns-1) = indexsphere(sphimage,-latp,longdbl);
      v(outcolumns+halfcolumns-xo+1,halfcolumns-yo+2) = indexsphere(sphimage,-latp,longdbr);

   end
end

function value = indexsphere(sphimage,latp,longd)
   rows=size(sphimage,1);
   columns=size(sphimage,2);
   xindex=round((longd+180)*(columns-1)/360+1);
   yindex=round((latp-90)*-(rows-1)/180+1);
   if longd <= 180 & longd >= -180 & latp <= 90 & latp >= -90
      value=sphimage(yindex,xindex);
   else
      value = 128;
   end
return
```

# Appendix B: *An In Situ Formula for the Peirce Quincuncial Projection*

In this appendix, we discuss the projection equations used in the Matlab source code provided in the previous appendix. Note that equations provided here is just a streamlined and simplified version of the pipeline shown on Figure 4. A simple formula for converting x and y coordinates in the Peirce square to latitude and longitude coordinates on the sphere in shown below. The equations are a consequence the fundamental mapping property of the Jacobi elliptic function cn(z,k) in the complex plane. A diagram for this is shown below the equations.

---

Peirce Quincuncial Projection: $(x, y) \to (\phi, \lambda)$
Input: x and y coordinates on the plane
  The values are confined to the square [-1,1] x [-1,1]
Output: latitude $\phi$ and longitude $\lambda$ on the sphere

$$w = cn(K_e(x - 1 + y\,i), \tfrac{1}{\sqrt{2}})$$

$$\lambda = \tan^{-1}\frac{Re(w)}{Im(w)}$$

$$\phi = (2\tan^{-1}\sqrt{(Re(w))^2 + (Im(w))^2} - \frac{\pi}{2})$$

---

Functions and constants used in the formula:
1) cn(z,k) – complex-valued Jacobi elliptic function
2) Re(z) – the real part of a complex variable
3) Im(z) – the imaginary part of a complex variable
4) $K_e$ is the complete Legendre elliptic integral of the 1$^{st}$ kind with parameter m = ½

$$K_e = K(\frac{1}{\sqrt{2}}) = \int_0^{\frac{\pi}{2}} \frac{dt}{\sqrt{1 - \frac{1}{2}\sin^2 t}} \approx 1.854$$

5) The intermediate variable $w$ is a complex variable with coordinate values in the unit disc

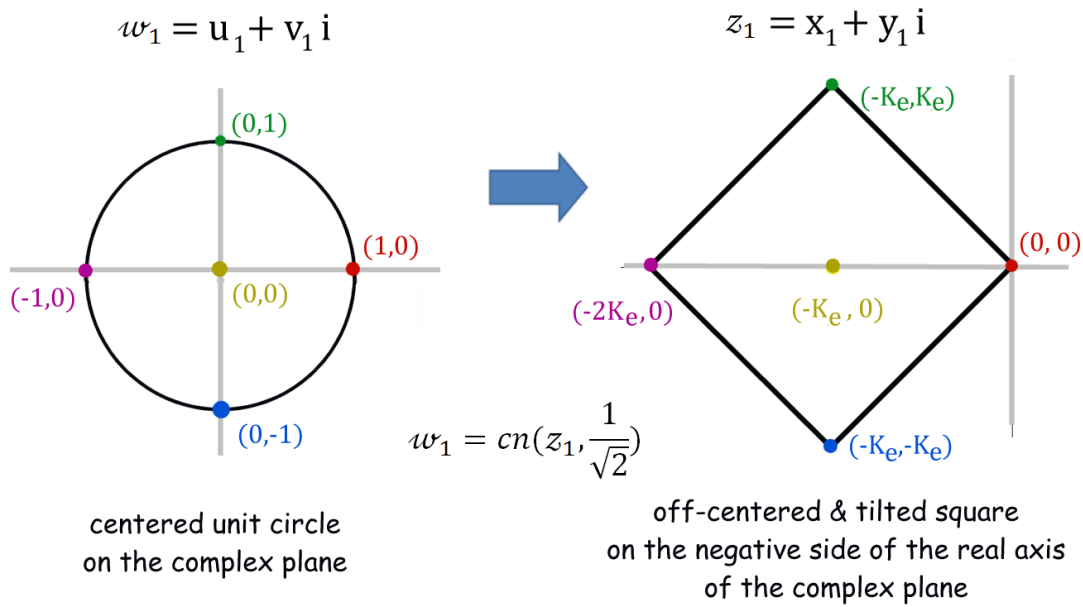

centered unit circle on the complex plane

$w_1 = cn(z_1, \tfrac{1}{\sqrt{2}})$

off-centered & tilted square on the negative side of the real axis of the complex plane

# Appendix C: ICIAM 2011

This paper was presented in the 7[th] International Congress of Industrial and Applied Mathematics. We include some of the visuals used in our presentation as part of the appendix.

## ICIAM logo

The ICIAM 2011 logo consists of a map of the world with a Canadian maple emblem over Vancouver. This world map almost looks like an equirectangular projection of the world. We decided to reproject this map to Peirce Quincuncial form for illustration purposes.

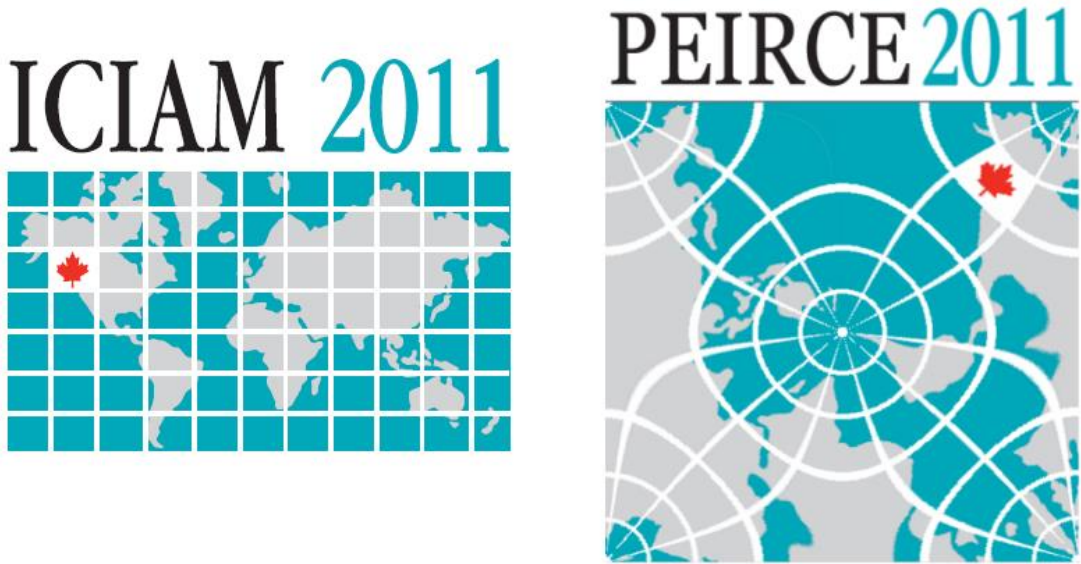

## A Peirce Quincuncial Orange

It is intuitive to think of performing map projections as analogous to peeling an orange. The orange can represent a sphere; and the orange peel can represent the surface of a sphere. This means that we can apply different map projections to an orange peel, including the Peirce quincuncial projection.

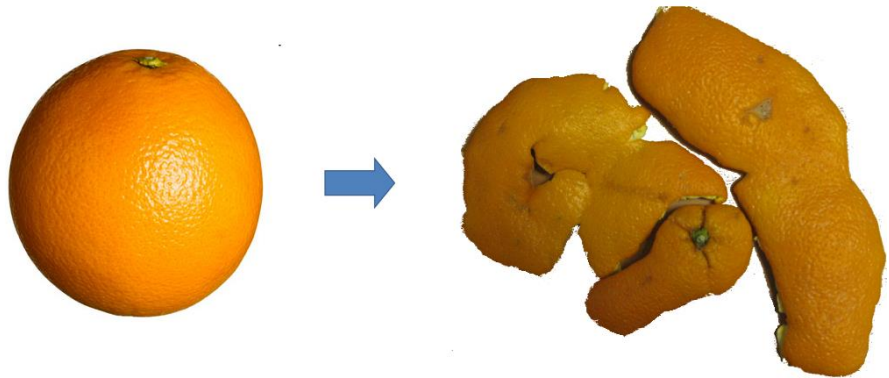

Map projections as peeling an orange

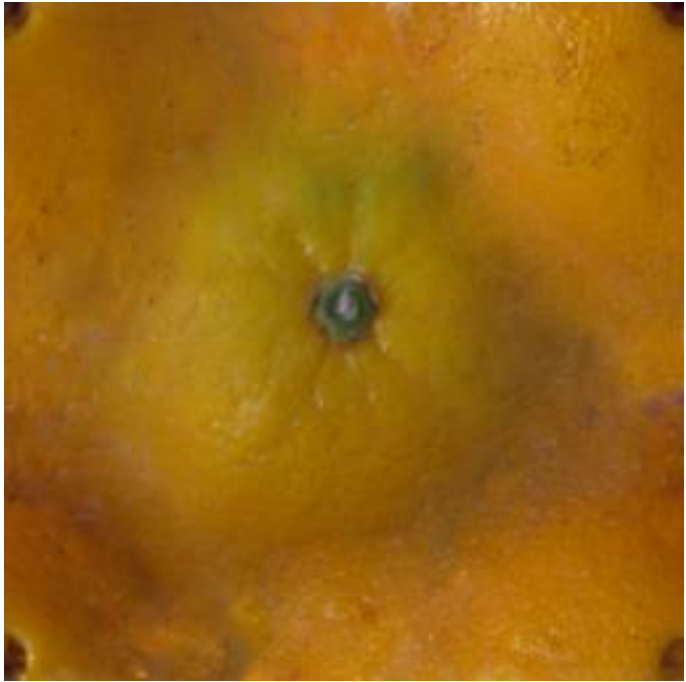

A Peirce quincuncial orange